\documentclass{article}
\usepackage[preprint]{arxiv_2023}
\usepackage[utf8]{inputenc} 
\usepackage[T1]{fontenc}    
\usepackage{hyperref}       
\usepackage{url}            
\usepackage{booktabs}       
\usepackage{amsfonts}       
\usepackage{nicefrac}       
\usepackage{microtype}      
\usepackage{xcolor}         

\usepackage{natbib}
\setcitestyle{numbers,open={[},close={]},citesep={,}} 
\usepackage{microtype}
\usepackage{graphicx}
\usepackage{subfigure}
\usepackage{booktabs} 
\usepackage{hyperref}

\usepackage{algorithm}
\usepackage{algorithmic}

\usepackage{amsmath}
\usepackage{amssymb}
\usepackage{mathtools}
\usepackage{amsthm}

\usepackage[capitalize,noabbrev]{cleveref}

\theoremstyle{plain}

\theoremstyle{definition}

\theoremstyle{remark}

\newcommand\thefont{\expandafter\string\the\font}
\usepackage{stmaryrd}

\usepackage{nicefrac}       
\usepackage{amsfonts,bm}
\usepackage{caption}
\usepackage{enumitem}

\title{Simple and Scalable Algorithms for \\ Cluster-Aware Precision Medicine}

\author{%
  Amanda M.~Buch
  \\
  Department of Psychiatry\\
  Weill Cornell Medicine\\
  New York, NY 10065 \\
  \texttt{amb2022@med.cornell.edu} \\
   \And
     Conor Liston \\
  Department of Psychiatry \\
  Weill Cornell Medicine \\
  New York, NY 10065 \\
  \texttt{col2004@med.cornell.edu} \\
   \And
  Logan Grosenick \\
  Department of Psychiatry \\
  Weill Cornell Medicine \\
  New York, NY 10065 \\
  \texttt{log4002@med.cornell.edu} \\
}

\begin{document}

\maketitle

\begin{abstract}
 AI-enabled precision medicine promises a transformational improvement in healthcare outcomes by enabling data-driven personalized diagnosis, prognosis, and treatment. However, the well-known ``curse of dimensionality'' and the clustered structure of biomedical data together interact to present a joint challenge in the high dimensional, limited observation precision medicine regime. To overcome both issues simultaneously we propose a simple and scalable approach to joint clustering and embedding that combines standard embedding methods with a convex clustering penalty in a modular way. This novel, cluster-aware embedding approach overcomes the complexity and limitations of current joint embedding and clustering methods, which we show with straightforward implementations of hierarchically clustered principal component analysis (PCA), locally linear embedding (LLE), and canonical correlation analysis (CCA). Through both numerical experiments and real-world examples, we demonstrate that our approach outperforms traditional and contemporary clustering methods on highly underdetermined problems (e.g., with just tens of observations) as well as on large sample datasets. Importantly, our approach does not require the user to choose the desired number of clusters, but instead yields interpretable dendrograms of hierarchically clustered embeddings. Thus our approach improves significantly on existing methods for identifying patient subgroups in multiomics and neuroimaging data, enabling scalable and interpretable biomarkers for precision medicine.
\end{abstract}

\section{Introduction}
\label{introduction}
In modern medicine, interpretable clustering of patients into distinct subtypes is increasingly important for personalized biomarker discovery, diagnosis, prognosis, and treatment selection \citep{Bishop2022-gj,Bonacchi2020-hz,Buch2023-hw,Drysdale2017-il,Santos2015-pb,Sorlie10869,Singh2018-sg,Qian2019-qb}. To facilitate adoption by healthcare professionals, we need explainable models that can be trained even when only limited data is available. However, due to the ``curse of dimensionality'', similarity metrics (and thus clustering algorithm outcomes) degrade in high dimensions (the “$p>N$” setting common in medical imaging, genomics, and multiomics, where we have $p$ correlated variables and $N$ observations fewer than $p$). As a result, it is popular to use a two-stage procedure where high dimensional data are first embedded into a low-rank representation, and then clustered in the resulting latent space. The mapping to the low-rank space (e.g., component loadings) are then often used to explain which variables are important (e.g., which differences in brain regions or genes relate to cluster differences \citep{Ciortan_undated-wh,Danda2021-qt,Drysdale2017-il,Gharavi2021-ms}).

\begin{figure*}[!t]
\centering\noindent
\resizebox{.87\columnwidth}{!}{
\includegraphics[scale=1]{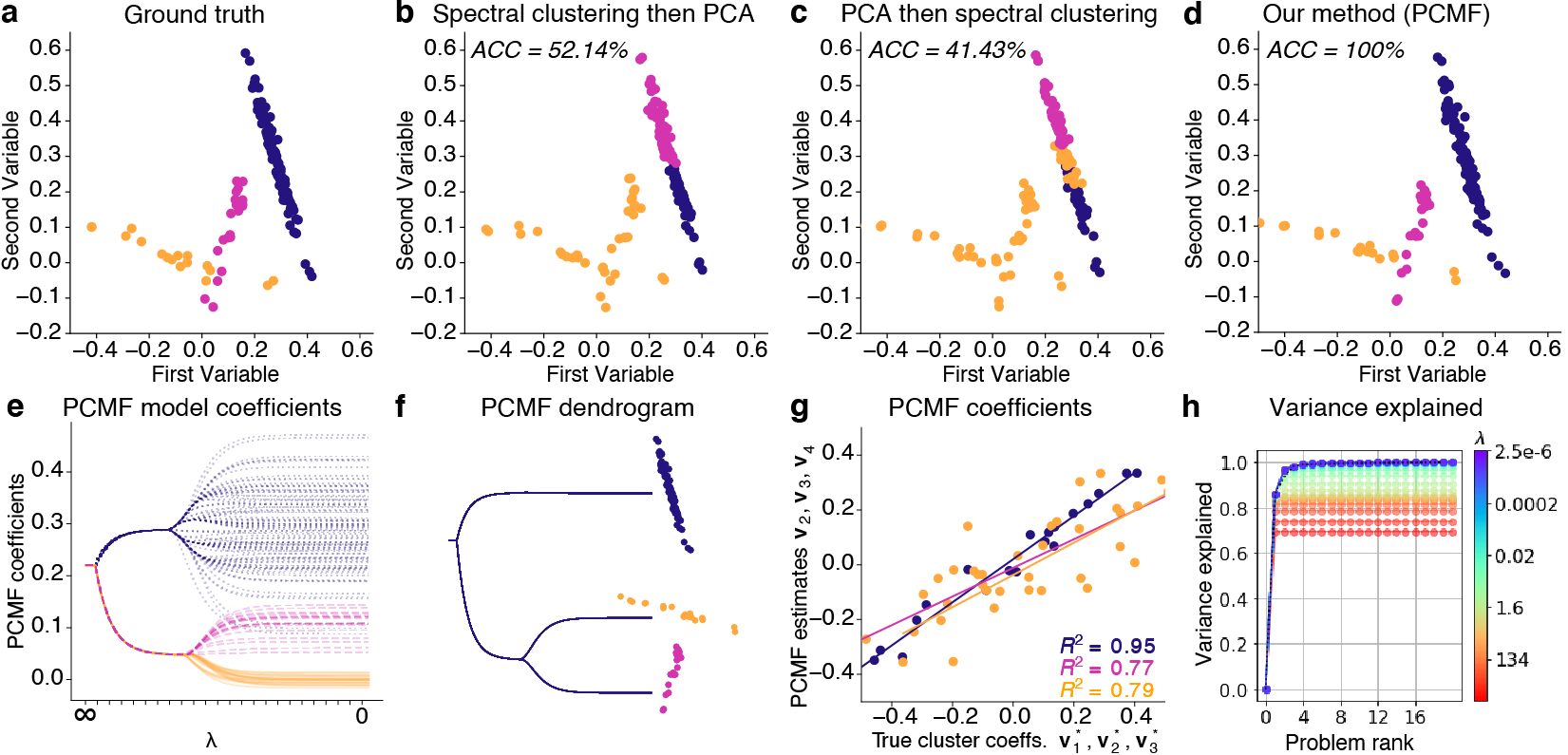}
}
\caption{PCMF for explainable joint PCA and hierarchical clustering. \textbf{a.} Scatterplot of reconstructed ground truth data (PCA rank $r=4$) for 3-class problem; $p=20; N_1=100$ (blue), $N_2=25$ (red), $N_3=25$ (orange), colored by true cluster membership. \textbf{b.} Spectral clustering sequentially followed by PCA ($r=4$) on clustered data. \textbf{c.} PCA ($r=4$) sequentially followed by spectral clustering on PCA components. \textbf{d.} Joint PCA and clustering with PCMF ($r=4; \lambda=3.0$). Two-step procedures in \textbf{b-c} fail to find correct clusters while PCMF succeeds. \textbf{b-d.} Color indicates predicted clusters. \textbf{e.} PCMF paths for variable 1 fit along decreasing penalty path ($\lambda=\infty$ to $\lambda=0$). \textbf{f.} Interpretable PCMF dendrogram estimated from paths. \textbf{g.} PCMF coefficients accurately fit ground truth cluster-specific coefficients used to generate data. PCMF coefficients $\mathbf{v_2}$, $\mathbf{v_3}$, and $\mathbf{v_4}$ approximate true cluster coefficients ("slopes") $\mathbf{v_1^*}$ (blue), $\mathbf{v_2^*}$ (red), and $\mathbf{v_3^*}$ (orange); $\mathbf{v_1}$ corresponds to cluster means intercept vector (not shown). \textbf{h.} Calculating variance explained by each PCMF component shows the rank $r=4$ model correctly captures the 3 cluster slopes and 1D-direction along which cluster means vary.}
\label{fig-1}
\vspace{-1.2em}
\end{figure*}

Unfortunately, such two-stage procedures can lead to suboptimal and hard-to-explain results  \citep{Chang1983-ai}, as the embedding ignores important clustered structure in the data, thereby harming the embedding. Further, as the embedding is agnostic to the underlying clusters, it may not provide a good space in which to separate the clusters (see Fig. \ref{fig-1}). These issues motivate a need for joint clustering and embedding methods for such data. Identical concerns extend to multiple datasets (“multiview" problems), where clustering and embedding has also typically been approached as a two-stage process; a low-rank, multiview embedding is obtained first and then input into a clustering algorithm (e.g., canonical correlation analysis (CCA) followed by clustering:  \citep{Buch2023-hw,Chen2008-uk, Chen2013-wl,Drysdale2017-il,Du2017-tt,Ouyang2019-rd}).

Recently, exciting new methods have emerged for jointly clustering and embedding data, including cluster-aware feature selection \citep{Wang2021-ql}, CCA mixture models \citep{Fern2005-jn,Lei2017-br}, non-negative matrix factorization (NMF)-based models \citep{Fogel2016-sa,Wu2020-bu,Zhou2021-wl}, and a number of neural networks (e.g, \citep{Boubekki2021-qk,Huang2014-tu,Lakkis2021-iz,Mautz2020-io,Shin2020-cw,Wang2016-ws,Yang2016-ak}. Although pioneering, these existing approaches involve complicated many-objective or deep neural network formulations that prioritize clustering over interpretability and have limited performance in restricted data cases, so far limiting their adoption in practice. 

Here, to develop an explainable and scalable formulation for joint clustering and embedding relevant to precision medicine applications, we show that a straightforward addition of a convex clustering penalty to standard embedding methods yields a simple, theoretically tractable, and modular approach to joint clustering and embedding that is highly competitive in practice and enjoys theoretical benefits over convex clustering in the ``large dimensional limit'' (LDL) regime appropriate for $p > N$ data (in the LDL regime $p/N \rightarrow c$ for constant $c$ as $p,N\rightarrow \infty$) \citep{Aparicio2020-dp,Bao2022-ci,Benaych-Georges2009-em,Couillet2022-yi,Dobriban2017-zo,Johnstone2001-zq,Paul2007-be}. 

\textbf{Main contributions and significance for precision medicine:}
\vspace{-1em}
\begin{enumerate}[leftmargin=*]
\setlength{\itemsep}{0pt}
\setlength{\parskip}{0pt}
\setlength{\parsep}{0pt}
\item We introduce a modular cluster-aware embedding strategy appropriate for precision medicine applications along with three fast and scalable algorithms that solve linear, locally linear, and multiview instantiations of this joint clustering and embedding approach.
\item We prove that our approach dominates convex clustering in the LDL regime.
\item Our approach does not require specifying cluster number. Instead it outputs \textit{interpretable} per-cluster embeddings organized as a dendrogram.
\item Our approach performs competitively against state-of-the-art methods on 14 real-world datasets.
\end{enumerate}
\vspace{-.2em}
\section{Related work: Convex clustering}
Clustering algorithms are classically formalized as discrete optimization problems that are NP-hard. However, by relaxing the hard clustering constraint to a convex penalty \citep{Pelckmans2005-re}, clustering can be reformulated as a convex optimization problem. In such “convex clustering”---also referred to as “clusterpath” or “sum-of-norms” clustering---the fitting procedure trades off approximating the data well with minimizing the sum of between-observation distances via a tuning penalty parameter, $\lambda$. The number of clusters is indirectly controlled by $\lambda$, and when solved along a path of $\lambda$s, convex clustering can exactly recover true data partitions among a mixture of Gaussians \citep{Hocking2011-pa,Jiang2020-gg,Lindsten2011-le}, and has been shown to converge \citep{Chi2019-ki,Radchenko2017-sp}. Further, the solution path can be visualized as a dendrogram to reveal hierarchical structure among clusters \cite{Weylandt2020-yc}.
More explicitly, for data matrix $X\in \mathbb{R}^{N \times p}$ with $N$ observations in the rows and $p$ variables in the columns, convex clustering solves the problem:
\begin{equation}
    \underset{\widehat{X} \in \mathbb{R}^{N \times p}}{\text{minimize}}\ \frac{1}{2}\|X-\widehat{X} \|_{F}^{2}+\lambda\sum_{i<j}w_{ij}\|\widehat{X}_{i\cdot}-\widehat{X}_{j\cdot}\|_{q}.
    \label{eq:1}
\end{equation}
Tuning $\lambda$ thus trades off the fidelity of a model-to-data fit term with a convex clustering penalty (which comes from a Lagrangian relaxation of an inequality constraint on the sum of the convex $q$-norms of differences between observations---typically $q\in \{1,2,\infty\}$). Importantly, weights $w_{i j} > 0$ constrained to be nonzero for nearest neighbors \citep{Chi2015-kb,Wang2021-ql} can speed up optimization and increase flexibility in modeling local structure in the row differences, such as with a radial basis function ($w_{i j} = \exp(-\gamma\|X_{i\cdot}-X_{j\cdot} \|_2^2)$ \citep{Chi2015-kb,Hocking2011-pa}, multiplicative weights \citep{Jiang2020-gg}, or properly scaling kernels \citep{Fodor2022-ca}. 

Recently there has been a flurry of theoretical and algorithmic developments for convex clustering \citep{Chiquet2017-gp,Fodor2022-ca,Jiang2020-gg,Lin2021-xh,Panahi2017-yp,Sui2018-ho,Sun2021-gf,Tan2015-js,Ieee2019-nn}, improving practically and theoretically on the number of  approaches that have been developed to solve problem \eqref{eq:1} \citep{Chi2015-kb,Hocking2011-pa, Panahi2017-yp, Sun2021-gf, Weylandt2020-yc}. Crucially, a warm-started ADMM approach---Algorithmic Regularization---was recently introduced to enable feasible computation of dense convex clustering $\lambda$ paths, speeding convergence more than $100$-fold \citep{Weylandt2020-yc}. Multiple studies have extended convex clustering, leading to new approaches to biclustering \citep{Allen2014-gb,Chi2015-kb}, multiview clustering \citep{Wang2021-ql}, and supervised convex clustering \citep{Wang2020-qq}. However, none of the existing approaches allow the same variables to contribute differently to multiple clusters, and none of them use the convex clustering penalty for joint clustering and embedding.

\section{Our approach: Pathwise Clustered Matrix Factorization (PCMF)}

\subsection{Pathwise Clustered Matrix Factorization (PCMF) problem formulation}
We propose using the convex clustering penalty as a modular addition to common embedding methods to make them cluster-aware (that is, to enable them to jointly cluster and embed). More explicitly, given a data matrix $X\in\mathbb{R}^{N\times p}$ (with $N$ observations in the rows, $p$ variables in the columns, and rank $R \leq \min(N,p)$), an embedding of $X$: $\widehat{X}\in\mathcal{M}_r$ (where $\mathcal{M}_r$ is a rank-$r$ manifold), and a loss function $\mathcal{L}(\cdot,\cdot):\mathbb{R}^{N\times p}\times \mathbb{R}^{N\times p} \rightarrow \mathbb{R}_+$, we can express our general problem as:
\begin{equation}
    \begin{aligned}
        &\underset{\widehat{X} \in \mathcal{M}_r}{\text{minimize}}\  \mathcal{L}(X,\widehat{X})+\lambda\sum_{i<j}w_{ij}\|\widehat{X}_{i\cdot}-\widehat{X}_{j\cdot}\|_{q},
    \end{aligned}
\label{eq:2}
\end{equation}
where the penalty term is identical to that used in convex clustering above but now applied to a jointly embedded $\widehat{X}\in \mathcal{M}_r$. To demonstrate the utility of this strategy, we begin with among the most well-known and widely-employed embedding algorithms: the truncated singular value decomposition (tSVD) \citep{Eckart1936-et}. In this case, expressing the embedding constraint $\widehat{X}\in \mathcal{M}_r$ explicitly in terms of the tSVD, equation \eqref{eq:2} yields the PCMF problem: 
\begin{equation}\tag{3}
    \begin{split}
    &\underset{
    \widehat{X}, U_r, S_r, V_r}{\text{minimize}}\ \frac{1}{2}\|X-\widehat{X}\|_{F}^{2}+\lambda\sum_{i<j}w_{ij}\|\widehat{X}_{i\cdot}-\widehat{X}_{j\cdot}\|_{q} \\[-0.5em] 
    &\text{subject to\ } \; \widehat{X}-U_rS_rV_r^{T}=0,\; U_r^{T}U_r=V_r^{T}V_r=I_{r},\; 
    S_r=\text{diag}(s_{1},\ldots,s_{r}),
    \end{split}\raisetag{12ex}
\label{eq:3}
\end{equation}
\addtocounter{equation}{1}for $s_{1}\geq s_{2}\geq\cdots\geq s_{r}>0$. Here the rank-$r \leq R$ tSVD embedding is given by $\widehat{X}=U_rS_rV_r^{T}$, subject to the usual orthogonality constraints on the first $r$ left and right singular vectors (collected in $U_r$ and $V_r$, respectively) and the standard ordering of the first $r$ singular values on the diagonal of $S_r$ \citep{Eckart1936-et}. Without loss of generality, we assume $X$ has been centered---a case where the tSVD is also called principal components analysis (PCA) (see Appendix $\S B.1.2$ for further discussion of uncentered effects). Note that when $r=R$ (that is, if $\text{rank}(\widehat{X}) = \text{rank}(X)$), this problem reduces to the standard convex clustering problem \eqref{eq:1} as a special case. We next present efficient algorithmic approaches to solving this nonconvex problem.

\subsection{Solving PCMF with the Alternating Direction Method of Multipliers (ADMM)}
Because in most cases it is desirable for many weights $w_{ij}$ in the convex clustering penalty to be exactly zero \citep{Chi2015-kb}, we first re-represent the relevant nonzero distances more efficiently as a sparse graph, ${G}$. We then introduce an auxiliary variable $G=D\widehat{X}\in \mathcal{R}^{|\mathcal {E}|\times p}$, where $D\in \mathbb{R}^{|\mathcal {E}|\times n}$ is a sparse matrix containing the weighted pairwise distances defined by edges $\mathcal{E}$. This allows us to rewrite the PCMF problem as:
\begin{equation}
    \begin{aligned}
        &\underset{\widehat{X},G,U_r,S_r,V_r}{\text{minimize}}\  \frac{1}{2}\|X-\widehat{X}\|_{F}^{2}+\lambda\sum_{\ell\in\mathcal{E}}w_{\ell}\|G_{\ell\cdot}\|_{q}\\[-0.25em]
        &\text{subject to\ } \widehat{X}-U_rS_rV_r^{T}=0,\; G-D\widehat{X}=0, &U^{T}U=V^{T}V=I_{r},\; S_r=\text{diag}(s_{1},\ldots,s_{r}),
    \end{aligned}
\label{eq:5}
\end{equation}for $s_{1}\geq\cdots\geq s_{r}>0$, which yields a problem separable in its objective and penalty subject to (nonconvex) constraints---a common application for ADMM. Algorithm \ref{alg:PCMF-ADMM} shows the resulting ADMM updates. Critically, we have added Algorithmic Regularization \citep{Weylandt2020-yc} along the $\lambda$ path. ADMM solutions fit along a path of $\lambda$s benefit from “warm-starting” by initializing the next problem along the path at the previous solution. Algorithmic Regularization takes this to the extreme, shortening steps along the path and decreasing the number of ADMM iterations at each point to a small number (achieved by making $K$ small in Algorithm \ref{alg:PCMF-ADMM}). For an appropriately chosen step size, this has been proven to converge to the true path solutions and to speed up the computation of path estimation by $>100$-fold \citep{Weylandt2020-yc}. This significantly improves computational feasibility as our algorithm requires solving over many path penalty ($\lambda$) values (see Appendix $\S C, D$ for derivation, convergence details, computational complexity, and consensus algorithm).

\begin{algorithm}
\caption{PCMF}\label{alg:PCMF-ADMM}
\begin{algorithmic}[1]
\setlength{\itemindent}{1em}
\item[\textbf{Input:} data $X$,  decreasing path $\{\lambda\}$, weights $\mathbf{w}$, pairwise distance matrix $D$]
\item[\textbf{Notation:} data mean $\overline{X}$, rank $r$, iteration $k$, norm $q\in\{1,2,\infty\}$, $\rho \geq 1$, operator $\text{prox}_{\frac{\lambda}{\rho}P_{\mathbf{w},q}(G)}$]
\setlength{\itemindent}{0em}
\STATE $G^0 \gets Z^0_1 \gets DX $; 
$\widehat{X} \gets Z^0_2 \gets \overline{X}$, $(U^0_r,S^0_r,V^0_r) \gets \text{SVD}_r(\widehat{X)}$, $L = \text{chol}(I + \rho I + \rho D^TD)$
\FOR{$\lambda \in \{ \lambda\}$}
\FOR{$k=1,\ldots,K$}
\STATE $\widehat{X}^{k+1} \gets L^{-T}L^{-1}\big(X+\rho D^T(G^k - Z_1^k) 
+\rho(U_r^kS_r^kV_r^{kT}-Z_2^k)\big)$
\STATE $G^{k+1} \gets \text{prox}_{\frac{\lambda}{\rho}P_{\mathbf{w},q}(G)}(D\widehat{X}^{k+1} + Z_1^k)$
\STATE $(U_r^{k+1},S_r^{k+1},V_r^{k+1}) \gets \text{SVD}_r(\widehat{X}^{k+1}+Z_2^k)$
\STATE $Z_1^{k+1} \gets Z_1^k + D^T\widehat{X}^{k+1} - G^{k+1}$
\STATE $Z_2^{k+1} \gets Z_2^k + \widehat{X}^{k+1} - U_r^{k+1},S_r^{k+1},V_r^{k+1}$
\ENDFOR
\STATE Save current path solutions: $\widehat{X}_\lambda \gets \widehat{X}^K$, $G_\lambda \gets G^K$, $(U_{r,\lambda}, S_{r,\lambda},V_{r,\lambda}) \gets (U_r^{K},S_r^{K},V_r^{K})$
\STATE Initialize for next path solution: $\widehat{X}^0 \gets \widehat{X}^K$, $G^0 \gets G^K$, $(U_{r}^0, S_{r}^0,V_{r}^0) \gets (U_r^{K},S_r^{K},V_r^{K})$
\ENDFOR
\STATE \textbf{return pathwise solutions} $\{\widehat{X}_\lambda\}, \{G_\lambda\}, \{U_{r,\lambda}\}, \{S_{r,\lambda}\},\{V_{r,\lambda}\}$
\end{algorithmic}
\end{algorithm}

\subsection{A nonlinear extension: locally linear PCMF (LL-PCMF)}
Next, we introduce a locally linear PCMF problem and a Penalized Alternating Least Squares (PALS) algorithm to solve it \citep{Roweis2000-ga}. For clarity (and without loss of generality), we center and scale $X$, set $s_1=1$, and consider the rank-$1$ version of the PCMF problem (which can be generalized to rank-$r$ using an appropriate deflation approach; see \citet{mackey2008deflation} and \citet{Witten2009-fs}). Then denoting the $i$th column vector of $X^T$ as $\mathbf{x}_{i}=(X^T)_{\cdot i}$ and defining penalty 
$\tilde{P}_{\mathbf{w},q}(\mathbf{u},\mathbf{v})=\sum_{(i,j)\in\mathcal{E}}w_{ij}\|u_{i}\mathbf{v}-u_{j}\mathbf{v}\|_{q}$, we can write the rank-$1$ tSVD with a convex clustering penalty (see Appendix $\S C.2$) as:
\begin{equation}
    \begin{split}
         & \underset{s,\mathbf{u},\mathbf{v}}{\text{minimize}}\ \sum_{i=1}^{N}\frac{1}{2}\|\mathbf{x}_{i}-u_{i}\mathbf{v}\|_{2}^{2}+\lambda \tilde{P}_{\mathbf{w},q}(\mathbf{u},\mathbf{v})\; 
        \text{subject to}\; \|\mathbf{u}\|_{2}^{2}=1,\ \|\mathbf{v}\|_{2}^{2}=1.
    \end{split}
\label{eq:6}
\end{equation}
To introduce the overparameterization necessary for convex clustering we replace the single vector $\mathbf{v}$ with a matrix $V\in \mathbb{R}^{p \times N}$ with column vectors $\mathbf{v}_i=V_{\cdot i}$ (denoting the set of these column vectors as $\{\mathbf{v}\}_i,\ i=1,\ldots,N$)---this allows each observation to  potentially be its own cluster in the limit $\lambda \rightarrow 0$. Note this is the same overparameterization as in the standard convex clustering problem \eqref{eq:1}. Defining $P_{\mathbf{w},q}(\mathbf{u},V)=\sum_{(i,j)\in\mathcal{E}}w_{ij}\|u_{i}\mathbf{v}_i-u_{j}\mathbf{v}_j\|_{q}$, we arrive at the overparameterized problem: \begin{equation}\tag{6}
    \begin{split}
        &\underset{\mathbf{u},V}{\text{minimize}}\   \sum_{i=1}^{N}\frac{1}{2}\|\mathbf{x}_{i}-u_{i}\mathbf{v}_{i}\|_{2}^{2}+\lambda P_{\mathbf{w},q}(\mathbf{u},V)\;
       \text{subject to}\; \|\mathbf{u}\|_{2}^{2}=1,\ \|\mathbf{v}_{i}\|_{2}^{2}=1, \ i=1,\ldots,N.
    \end{split}\raisetag{8ex}\
\label{eq:7}
\end{equation}
 Next, by removing the cross-terms in the penalty we allow $\mathbf{u}$ and $\mathbf{v}$ to independently vary and the locally-defined weights $w_{ij}$ to apply to the embedding (making it locally linear). We do this by replacing $P_{\mathbf{w},q}(\mathbf{u},\mathbf{v})$ with $Q^\mathbf{u}_{\mathbf{w},q}(\mathbf{u})=\sum_{(i,j)\in\mathcal{E}}w_{ij}|u_i-u_j|$ and $Q^V_{\mathbf{w},q}(V)=\sum_{(i,j)\in\mathcal{E}}w_{ij}\|\mathbf{v}_i-\mathbf{v}_j\|_{q}$. Then using fixed values from iterate $k$, $y_{\mathbf{u},i}^{k}=\mathbf{x}_{i}^T\mathbf{v}_{i}^{k}$ and
$\mathbf{y}_{\mathbf{v},i}^{k}=u_{i}^{k}\mathbf{x}_{i}$, we get updates:
\begin{equation}\tag{7a}
    \begin{split}
    &\mathbf{u}^{k+1} \leftarrow  \underset{\mathbf{u}}{\text{argmin}}\ \sum_{i=1}^{N}\|y_{\mathbf{u},i}^{k}-u_{i}\|_{2}^{2}+\lambda Q^\mathbf{u}_{\mathbf{w},q}(\mathbf{u})\; 
    \text{subject to\ }\|\mathbf{u}\|_{2}^{2}=1, i=1,\ldots,N,
    \end{split}
\label{eq:9a}
\end{equation}
\vspace{-4pt}
\begin{equation}\tag{7b}
    \begin{split}
    &\{\mathbf{v}_{i}\}{}^{k+1} \gets \underset{\{\mathbf{v}_{i}\}}{\text{argmin}}\ \sum_{i=1}^{N}\|\mathbf{y}_{\mathbf{v},i}^{k}-\mathbf{v}_{i}\|_{2}^{2}+\lambda Q^V_{\mathbf{w},q}(V)\; 
    \text{subject to\ }\|\mathbf{v}_{i}\|_{2}^{2}=1,\ i=1,\ldots,N.
    \end{split}
\label{eq:9b}
\end{equation}
\addtocounter{equation}{3}Note these iterative PALS updates for LL-PCMF are just convex clustering problems with constraints, and thus given some convex clustering solver \textsc{ConvexCluster} (e.g., $\S C.4$ in the Appendix), we arrive at our algorithm for LL-PCMF  (see Appendix $\S C$ for algorithm and derivation). 
\subsection{A multiview extension: Solving Pathwise Clustered CCA (P3CA) with PALS}
We next extend our approach to multiview learning, where we aim to jointly learn low-rank correlation structure while clustering observations across multiple data views (i.e., fitting canonical correlation analysis or CCA within clusters). To do so, we follow a derivation similar to LL-PCMF (note it is also straightforward to derive a linear P3CA by instead replacing the SVD with CCA in Alg. \ref{alg:PCMF-ADMM}), introducing the overparameterized pathwise clustered canonical correlation analysis (P3CA) optimization problem (recall $\mathbf{v}_i = V_{\cdot i}$ are column vectors of $V\in \mathbb{R}^{p \times N}$). We have data matrices $X\in\mathbb{R}^{N\times p_{X}},Y\in\mathbb{R}^{N\times p_{Y}}$ and variables $\mathbf{u}_{i}\in\mathbb{R}^{p_{X}},\ \mathbf{v}_{i}\in\mathbb{R}^{p_{Y}}$, and we define $\Sigma_i = X_{i\cdot}^TY_{i\cdot} \in \mathbb{R}^{p_X \times p_Y}$ and $Q_{\mathbf{w},q}(V)=\sum_{(i,j)\in\mathcal{E}}w_{ij}\|\mathbf{v}_i-\mathbf{v}_j\|_{q}$. This yields the penalized rank-1 CCA problem:
\begin{equation}\tag{8}
    \begin{split}
    &\underset{\{\mathbf{u}_{i}\},\{\mathbf{v}_{i}\}}{\text{maximize}}\sum_{i=1}^{N}\mathbf{u}_{i}^T\Sigma_i\mathbf{v}_{i}\ - \lambda Q_{\mathbf{w},q}(U) - \lambda Q_{\mathbf{w},q}(V)\; 
    \text{subject to} \ \|\mathbf{u}_{i}\|_{2}^{2}=1,\ \|\mathbf{v}_{i}\|_{2}^{2}=1, \\[-0.5em]
    \end{split}
\label{eq:10}
\end{equation}
for $i=1,\ldots,N$. Without inequality constraints, this is a biconvex problem in the $\{\mathbf{u}_{i}\}$ and $\{\mathbf{v}_{i}\}$ when the subproblems are relaxed by fixing $\tilde{\mathbf{x}}_{i}=\Sigma_i\mathbf{v}_{i}$
and $\tilde{\mathbf{y}}_{i}=\Sigma_i^{T}\mathbf{u}_{i}$
at each subiterate: 
\begin{equation}\tag{9a}
    \begin{split}
    &  \{\mathbf{u}_{i}\}^{k+1} \gets \underset{\{\mathbf{u}_{i}\}}{\text{argmin}}\sum_{i=1}^{N}\frac{1}{2}\|\tilde{\mathbf{x}}_{i}-\mathbf{u}_{i}\|_{2}^{2}+\lambda Q_{\mathbf{w},q}(U)\
   \text{subject to}\ \|\mathbf{u}_{i}\|_{2}^{2}=1,\ i=1,\ldots,N,
    \end{split}\raisetag{4ex}
\label{eq:11a}
\end{equation}
\vspace{-1em}
\begin{equation}\tag{9b}
    \begin{split}
    &\{\mathbf{v}_{i}\}^{k+1} \ \gets \underset{\{\mathbf{v}_{i}\}}{\text{argmin}}\sum_{i=1}^{N}\frac{1}{2}\|\tilde{\mathbf{y}}_{i}-\mathbf{v}_{i}\|_{2}^{2}+\lambda Q_{\mathbf{w},q}(V)\; 
    \text{subject to}\ \|\mathbf{v}_{i}\|_{2}^{2}=1,\ i=1,\ldots,N.
    \end{split}\raisetag{4ex}
\label{eq:11b}
\end{equation}
Each update is again a constrained convex clustering problem, leading to Algorithm \ref{alg:P3CA}. Empirically, for sufficiently small steps sizes, Algorithmic Regularization closely approaches the ADMM solutions with a significant speed up (see Appendix $\S C$ and $\S D.3$ for derivation and  computational complexity).

\begin{algorithm}[!htb]
\caption{Pathwise Clustered Canonical Correlation Analysis (P3CA)}\label{alg:P3CA}
\begin{algorithmic}[1]
\setlength{\itemindent}{1em}
\item[\textbf{Input:} data $(X,Y)$, decreasing path $\{\lambda\}$, weights $\mathbf{w}$, norm $q\in\{1,2,\infty\}$]
\item[\textbf{Notation:} iteration $k$, data means $(\bar{X},\bar{Y})$, $\mathbf{v}_i=V_{\cdot i}$, $\tilde{\mathbf{x}}_i=(\tilde{X}_{i \cdot})^T$, $\tilde{\mathbf{y}}_i=(\tilde{Y}_{i\cdot})^T$, $\rho \geq 1$]
\setlength{\itemindent}{1em}
\setlength{\itemindent}{0em}
\STATE $U \gets \bar{X}, V \gets \bar{Y}$
\FOR{$\lambda \in \{ \lambda\}$}
\FOR{$k=1,\ldots,K$}
\STATE $\tilde{\mathbf{x}}_{i}^{k+1} \gets \Sigma_i\mathbf{v}_{i}^{k} \ (\Sigma_i = X_{i\cdot}Y_{i\cdot}^T \in \mathbb{R}^{p_X \times p_Y})\ \text{for} \ i=1,\ldots,N$
\STATE $\mathbf{u}_i^{k+\frac{1}{2}} \gets \small{\textsc{ConvexCluster}}(\tilde{X}^{k+1}, U^k, \lambda, \mathbf{w}$, q) \\
\STATE $\mathbf{u}_i^{k+1} \gets \text{prox}_{\|\cdot\|_2^2}(\mathbf{u}_i^{k+\frac{1}{2}})\ \text{for} \ i=1,\ldots,N$
\STATE $\tilde{\mathbf{y}}_{i}^{k+1} \gets \Sigma_i^{T}\mathbf{u}_{i}^{k+1} \ (\Sigma_i^T = Y_{i\cdot}X_{i\cdot}^T \in \mathbb{R}^{p_Y \times p_X})\ \text{for} \ i=1,\ldots,N$
\STATE $\mathbf{v}_i^{k+\frac{1}{2}} \gets \small{\textsc{ConvexCluster}}(\tilde{Y}^{k+1}, V^k,  \lambda, \mathbf{w}$, q) \\
\STATE $\mathbf{v}_i^{k+1} \gets \text{prox}_{\|\cdot\|_2^2}(\mathbf{v}_i^{k+\frac{1}{2}})\ \text{for} \ i=1,\ldots,N$
\ENDFOR
\STATE Save path solutions: $U^K_{i \cdot} \gets \mathbf{u}^{KT}_i;\ V^K_{i \cdot} \gets \mathbf{v}^{KT}_i \ \text{for} \ i=1,\ldots,N $; $(U_\lambda, V_\lambda) \gets (U^K, V^K)$
\STATE Initialize: $(U^0, V^0) \gets (U^K, V^K)$
\ENDFOR
\STATE \textbf{return pathwise solutions} $\{U_\lambda\}, \{V_\lambda\}$
\end{algorithmic}
\end{algorithm}
\subsection{PCMF for Gaussian Mixture Model (GMM) data in the Large Dimensional Limit (LDL)}
Here we show that our approach dominates convex clustering in the $p > N$ LDL regime relevant to precision medicine, offering a constructive proof for inadmissibility of convex clustering in the case of ``nontrivially'' clustered GMM data using results from random matrix theory (RMT) \citep{Couillet2022-yi}.

\textbf{Definition 1} ``Nontrivial'' GMM data. We observe $N$ i.i.d. data vectors $\mathbf{x}_i \in \mathbb{R}^p$ drawn from the $K$-class GMM with fixed class sizes $N_1,\ldots,N_K$ (with $\sum_{k=1}^K N_k = N$) gathered in data matrix $X = [\mathbf{x}_1,\ldots,\mathbf{x}_N]^T \in \mathbb{R}^{N \times p}$, with $p \sim N$ or $p > N$ such that $p/N \rightarrow c\in (0,\infty)$ and  $N_a/N \rightarrow c_a\in (0,1)$ as $N,N_a,p\rightarrow \infty$. Letting $\mathcal{C}_a$ be the set of observations from class $a$ for $a \in \{1,\ldots,K\}$ such that $\mathbf{x}_i \sim \mathcal{N}(\mu_a,C_a) \iff \mathbf{x}_i \in \mathcal{C}_a$ with $C_1,\ldots,C_k$ distinct and of bounded norm. To ensure that cluster separation is nontrivial as $N,p \rightarrow \infty$ we take $\|\mathbf{x}_i\|$ to be of order $O(\sqrt{p})$ and $\|\mu_a - \mu_b\| = O(1)$ for $a,b \in 1,\ldots,K; a \neq b$. See Appendix $\S E$ for details and motivation.

\textbf{Proposition 1} For clustering the nontrivial GMM data in the LDL regime, PCMF asymptotically dominates standard convex clustering.

\textbf{Proposition 2} For clustering the nontrivial GMM data in the LDL regime, the local linear relaxation LL-PCMF asymptotically dominates convex clustering.

\textbf{Proposition 3} PCMF generalizes kernel spectral clustering (kSC) to joint clustering and embedding.

We briefly explain these results here leaving proofs to the Appendix ($\S E$). To prove Proposition 1 we let $C^\circ = \sum_{a=1}^k \frac{N_a}{N} C_a$ and note that due to ``universality'' results from RMT (see \citep{Couillet2022-yi} Ch. 2) the GMM assumption is often (though not always) equivalent to requiring $x_i \in \mathcal{C}_a : x_i = \mu_a + C_a^{1/2}z_i$ (where $z_i$ is a random vector with i.i.d. zero mean, unit variance, and suitably bounded higher-order moment entries) \citep{Couillet2022-yi}. We then consider the convex clustering penalty ``element-wise'' and find that:
\begin{align*}\tag{10}
\sum_{i,j \in \mathcal{E}} f\left(\frac{1}{p} \|\widehat{\mathbf{x}}_i - \widehat{\mathbf{x}}_j \|_2^2 \right) = \sum_{i,j \in \mathcal{E}} f\left(\frac{2}{p}C^\circ + O(p^{-1/2})\right),
\label{eq:RMT}
\end{align*}
where we have subsumed weights $w_{ij}$ and the square root into $f(\cdot)$ and normalized by $p$. The equality in \eqref{eq:RMT} follows from expanding and considering each term given the assumptions above, and indicates that if we consider the entry-wise distances in the LDL regime, all entries are dominated by the constant $O(1)$ term $2\text{tr}\ C^\circ /p$ regardless of the values of $a$ and $b$. Further $\|\mu_a - \mu_b\|_2^2/p=O(p^{-1})$ is dominated by the $O(p^{1/2})$ noise terms, indicating convex clustering's entry-wise distance approach does not allow discrimination for such data in this regime. 

However, by instead considering the data ``matrix-wise'' via embedding, we find that there is important discriminative information available in the low-rank structure of the Euclidean distances. In particular, although the matrix is again dominated by a $O(N)$-norm constant matrix $2\text{tr}\ C^\circ/p \cdot \mathbf{1}_N\mathbf{1}^T_N$, this rank-1 term can be discarded leaving resolvable spectral information about the covariance traces in the $O(N^{-1/2}$)-norm rank-2 second dominant term. Further, subsequent order $O(1)$ terms contain usable discriminative information about the means. Thus, using a low-rank embedding in the convex clustering term (PCMF) allows discrimination in the nontrivial GMM case and standard convex clustering does not. The proof of Proposition 2 follows similarly. Proposition 3 follows from noting that the differences in the penalty term can be represented through an application of the Laplacian of the weight-induced graph such that PCMF is jointly optimizing both a (kernel) spectral embedding and a clustering on that embedding (see Appendix $\S$E for proofs). Together, these propositions show PCMF outperforms convex clustering in $p > N$ problems and formalize its relationship to kSC.

\subsection{PCMF dendrograms for explainability and model selection}
PCMF fits a path of solutions along a sequence of values of $\lambda$ (Fig. \ref{fig-1}e--f), and when using the $\ell_2$-norm ($q=2$) (as we do below, given its desirable rotational symmetry), not all members of a cluster are shrunk to exactly the same value \citep{Hocking2011-pa}. Previous work has forced hard clustering at each agglomerative stage along the $\lambda$ path \citep{Hocking2011-pa,Weylandt2020-yc,Jiang2020-gg}. This may artificially force observations into one cluster that may then later switch to another, resulting in nonsmooth paths in practice. We choose to instead let the paths be unconstrained and smooth while solving divisively, and then to generate a dendrogram using a wrapper function that estimates sequential split points from the fully-fit paths by sequentially testing whether increasing the number of clusters at each step would improve overall model fit in terms of the penalized log-likelihood. Clustering at each $\lambda$ is performed on the weighted affinity matrix generated from differences matrix defined by the dual variables as recommended in \citep{Chi2015-kb}. Thus, this procedure estimates the connected components of the affinity graph defined by the dual variables at each value of $\lambda$. Further details on model selection are described in Appendix $\S F$.
\section{Experimental results}
\begin{figure}[!htb]
    \centering\noindent
    \resizebox{1\columnwidth}{!}{\includegraphics{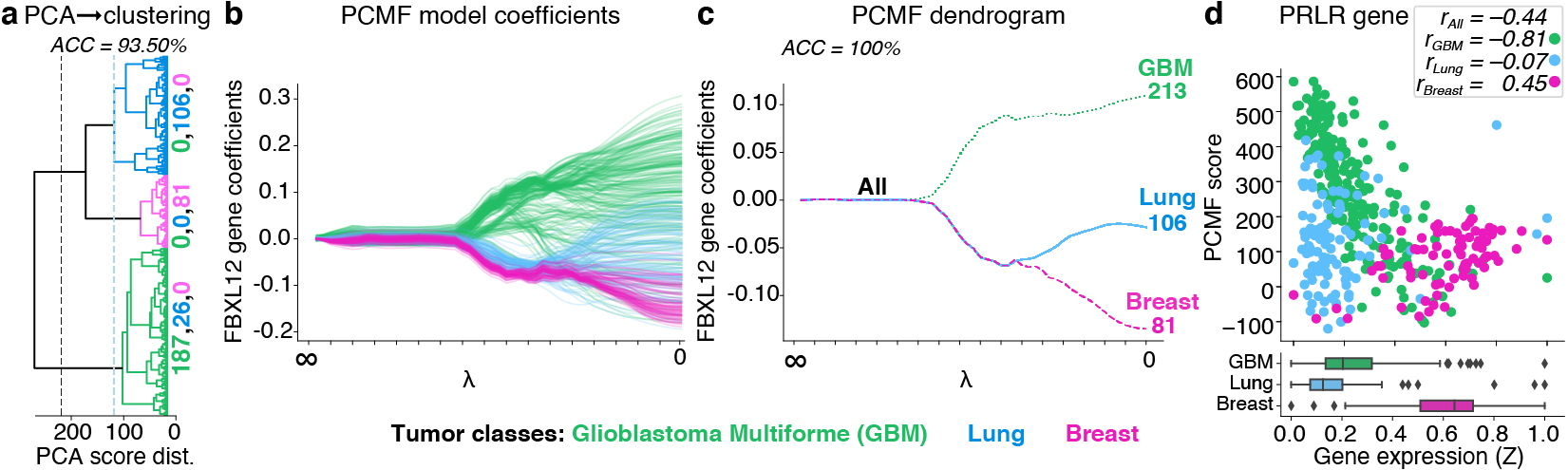}}
     \caption{PCMF identifies tumor clusters and embeddings using gene expression ($p=11,931$) from $N=400$ samples. \textbf{a.} Dendrogram shows hierarchical clustering on PCA embedding (PCA---> clustering). \textbf{b.} PCMF path and \textbf{c.} dendrogram shows PCMF perfectly recovers clusters ($ACC = 100\%$). \textbf{d.} Scatterplots and boxplots show distribution of PRLR gene expression versus PCMF expression scores for each sample colored by PCMF-predicted clusters. $r$: correlation between gene expression and PCMF expression score. Abbreviations: Accuracy, ACC; distance, dist.
    }
\label{tumor-Figure}
\end{figure}

\begin{figure}[!htb]
    \centering\noindent
    \resizebox{1\columnwidth}{!}{
    \includegraphics{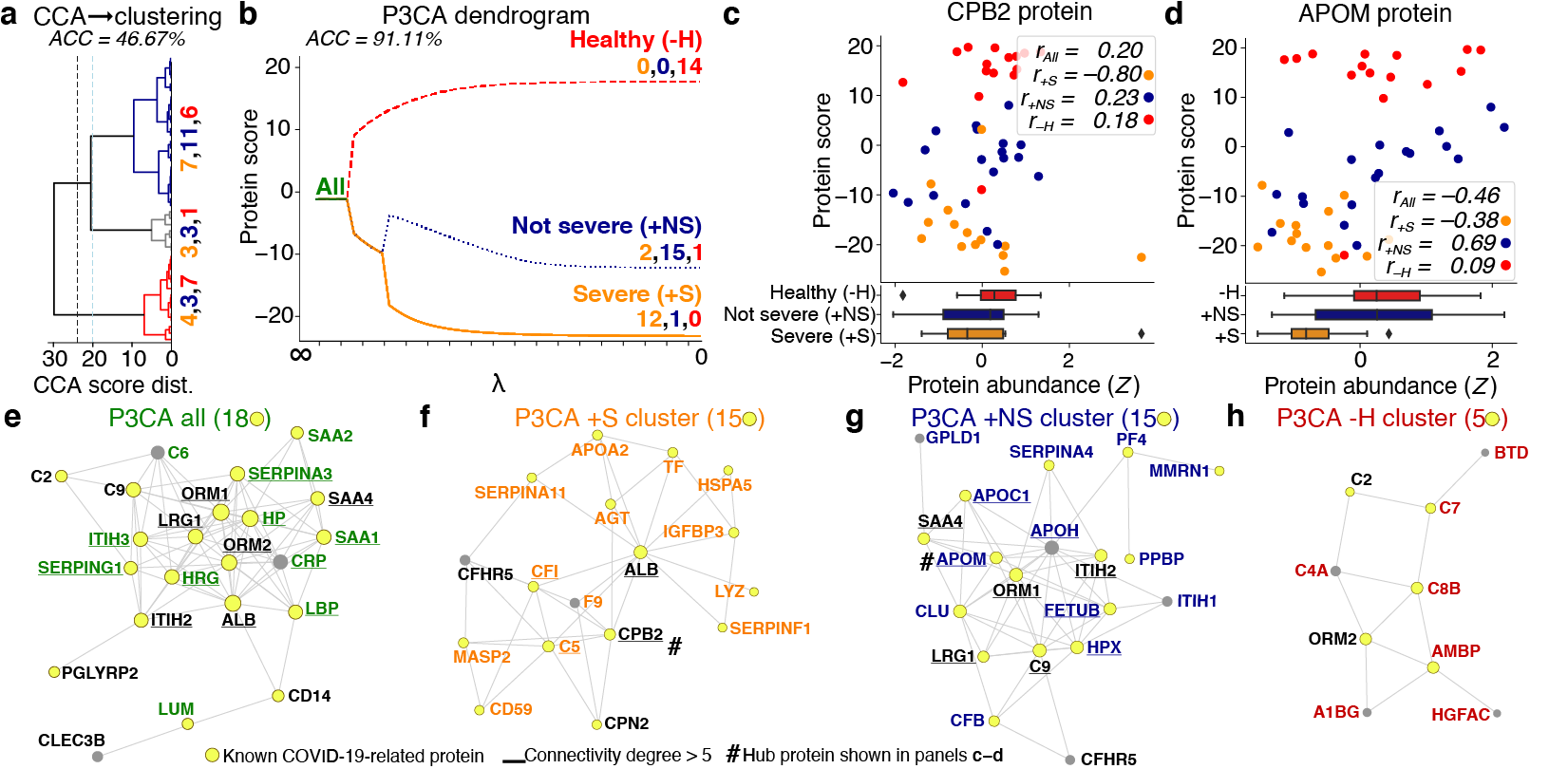}}
    \caption{P3CA identifies COVID-19-severity clusters and embeddings using protein ($p=382$) and metabolite ($p=403$) abundances from $N=45$ individuals. \textbf{a.} Dendrogram shows hierarchical clustering on CCA embedding (CCA---> clustering) fails to identify clusters. \textbf{b.} P3CA dendrogram shows P3CA accurately identifies clusters ($ACC = 91.11\%$). \textbf{c-d.} Scatterplots and boxplots show distribution of protein abundance versus P3CA protein scores for each individual colored by P3CA-predicted clusters. $r$: correlation between abundances and P3CA score. \textbf{e-h.} Protein-protein interaction (PPI) networks for top 25 proteins associated with each P3CA-predicted cluster. Colored text: cluster-specific proteins; yellow node: COVID-19-associated; underlined: network degree > 5
    }
\label{COVID-Figure}
\vspace{-1em}
\end{figure}

We evaluate our unsupervised PCMF methods in numerical experiments using 22 synthetic datasets (Table \ref{tab:Table-1}, Fig. 1, Appendix Tables 1--2, and Appendix Figs. 1--4,8--9) and on 14 biomedical datasets (Table \ref{tab:Table-1} and Appendix Figs. 5--7) using small underdetermined (limited observations $N$, $p>N$; Table \ref{tab:Table-2}, Appendix Tables 3--4, and Fig. 3) and large (many observations $N$; Table \ref{tab:Table-3} and Fig. 2) biomedical datasets. We measure clustering accuracy (ACC) using PCMF, LL-PCMF, and P3CA against 14 other clustering methods: (1) PCA/CCA+K-means \citep{Hastie2009-jq,Hotelling1933-ij,Hotelling1936-vz,MacQueen1967-ww}, (2) Ward \citep{Hastie2009-jq}, (3) spectral \citep{Hastie2009-jq}, (4) Elastic Subspace \citep{You2016-ir,You2016-ix}, (5) DP-GMM  \citep{You2016-ir}, (6) gMADD \citep{Paul2021-kl,Sarkar2020-le}, (7) HDCC \citep{Berge2012-pd,Bouveyron2007-ah}, (8) Leiden \citep{Traag2019-li}, (9) Louvain \citep{Blondel2008-tg}, (10) DP-GMM \citep{Escobar1995-pt}, (11) convex clustering (hCARP) \citep{Weylandt2020-yc}), (12) Deep Embedding Clustering (DEC) \citep{Xie2015-uq}, (13) IDEC \citep{Guo2017-cp}, and (14) CarDEC \citep{Lakkis2021-iz}. We visualize and interpret embedding cluster hierarchy in dendrograms of PCMF model solution paths (Figs. \ref{fig-1}--\ref{COVID-Figure}). 

\begin{table*}[!htb]
\centering
\caption{Datasets in Tables \ref{tab:Table-2}-\ref{tab:Table-3} and cluster discovery analysis. See Appendix $\S G$ for more details.}
    \resizebox{.89\columnwidth}{!}{%
    \begin{tabular}{rlcc} 
    \toprule
    Dataset & Variables ($p)$ & Samples ($N$) & Classes \\
    \midrule
    NCI & $6,830$ genes (expression) ($X$) & $64$  & $13$ cell types \\
    SRBCT & $2,318$ genes (expression) ($X$) & $88$  & $4$ cancer diagnoses\\
    Mouse & $16,944$ genes (scRNA-seq) ($X$) & $125$ & $7$ mouse organ types\\
    Tumors & $11,931$ expression/methylation ($X$) & $142$ & $3$ cancer diagnoses\\
    Tumors-Large & $11,931$ expression/methylation ($X$) & $400$ & $3$ cancer diagnoses\\
    MNIST & $784$ image pixels from 28 x 28 pixel image ($X$) & $36,000$ & $6$ digit types \\
    MNIST Fashion & $784$ image pixels from 28 x 28 pixel image ($X$) & $36,000$ & $6$ clothing types \\
    Synthetic data & $1,000$ synthetic variables; $\delta=0.5$ ($X$) & $100,000$ & $4$ clusters \\
    COVID-19 (Multiview) & $403$ metabolites ($X$); $382$ proteins ($Y$) & $45$ & $3$ severities \\
    NCI (Multiview) & $1,000$ genes (expression) ($X$); $100$ genes (expression) ($Y$) & $64$  & $13$ cell types \\
    SRBCT (Multiview)& $1,000$ genes (expression) ($X$); $100$ genes (expression) ($Y$) & $88$  & $4$ cancer diagnoses\\
    Mouse (Multiview)& $1,000$ genes (scRNA-seq) ($X$); $100$ genes (scRNA-seq) ($Y$)  & $125$ & $7$ mouse organ types\\
    Tumors (Multiview)& $1,000$ genes (expression) ($X$); $100$ genes (expression) ($Y$) & $142$ & $3$ cancer diagnoses\\
    Autism (ASD) (Multiview) & $3$ behaviors ($X$); $20$ RSFC features ($Y$) & $299$ & Unknown (discovery analysis) \\ 
   {Palmer Penguin (Multiview)} & $2$ features ($X$); $2$ features ($Y$) & $342$  & $3$ penguin species\\

    \bottomrule
    \end{tabular}
    }
\label{tab:Table-1}
\end{table*}

\textbf{Numerical experiments using PCMF demonstrate scalability via consensus formulation.} We find that PCMF and LL-PCMF with nearest neighbors $N.N. = 25$ performs competitively in accuracy in 12 synthetic datasets, especially for $ p> N$ ($p = 200$, $p = 2,000$; Appendix Table 1). We further show our consensus ADMM runs on larger $N$ datasets (unlike standard convex clustering, which cannot run on $N > 1,000$) demonstrating scalability to synthetic datasets of $N = 100,000; p = 1,000$ (Appendix Fig. 3 and Appendix Table 2) and $N = 100,000; p = 1,000$, and has high cluster assignment accuracy in held-out test set data (Table \ref{tab:Table-3}). To evaluate the PCA interpretation of PCMF embeddings, we compare and show high similarity to tSVD estimates fit within ground-truth clusters (Fig. \ref{fig-1}g, Appendix $\S B.1.2$, Appendix Table 2, and Appendix Figs. 3--4).

\begin{table*}[!htb]
\centering
\caption{Clustering accuracy on small real-world datasets (``MV'' abbreviates ``Multiview'').}
\resizebox{.88\columnwidth}{!}{%
\begingroup
\setlength{\tabcolsep}{4pt}
\begin{tabular}{lcccccccccc}
\toprule
& \normalsize{NCI} & \normalsize{SRBCT} & \normalsize{Mouse} & \normalsize{Tumors} & \normalsize{COVID-19} & \normalsize{Penguins} & \normalsize{NCI-MV} & \normalsize{SRBCT-MV} & \normalsize{Mouse-MV} & \normalsize{Tumors-MV}\\

\color[HTML]{F7890D} \normalsize{\textbf{PCMF}} &
{\color[HTML]{F7890D} \normalsize{43.79\%}} &
{\color[HTML]{F7890D} \normalsize{51.8\%}} &
{\color[HTML]{F7890D} \normalsize{73.6\%}} &
{\color[HTML]{F7890D} \normalsize{92.25\%}} &
{—} &
{—} &
{—} &
{—} &
{—} &
{—} \\

\color[HTML]{F7890D} \normalsize{\textbf{LL-PCMF}} &
\textbf{{\color[HTML]{F7890D} \normalsize{64.06\%} }} &
\color[HTML]{F7890D} \normalsize{55.42\%} &
\textbf{{\color[HTML]{F7890D} \normalsize{80.00\%} }} &
\textbf{{\color[HTML]{F7890D} \normalsize{97.89\%} }} &
{—} &
{—} &
{—} &
{—} &
{—} &
{—} \\

{\color[HTML]{F7890D} \normalsize{\textbf{P3CA}}} &
{—} &
{—} &
{—} &
{—} &
{\color[HTML]{F7890D} \normalsize{\textbf{91.11}\%}} &
{\color[HTML]{F7890D}\normalsize{\textbf{98.25\%}}} &
{\color[HTML]{F7890D}\normalsize{\textbf{56.25\%}}} &
{\color[HTML]{F7890D}\normalsize{\textbf{65.06}\%}} &
{\color[HTML]{F7890D}\normalsize{\textbf{63.20\%}}} &
{\color[HTML]{F7890D}\normalsize{\textbf{98.59\%}}} \\

\normalsize{PCA + K-means} & \normalsize{39.06\%} & \normalsize{40.96\%} & \normalsize{45.60\%} & \normalsize{50.00\%} & — & — & — & — & — & — \\

\normalsize{CCA + K-means} & — & — & — & — & \normalsize{51.11\%}   & {\color[HTML]{000000} \normalsize{79.82\%}} &
\normalsize{31.25\%} &
\normalsize{37.35\%} &
\normalsize{27.20\%} &
\normalsize{50.70\%} \\

\normalsize{Ward} & \normalsize{56.25\%} & \normalsize{40.96\%} & \normalsize{46.40\%} & \normalsize{94.37\%} & \normalsize{68.89\%}  & {\color[HTML]{000000} \normalsize{96.78\%}} &
\normalsize{51.56\%} &
\normalsize{40.96\%} &
\normalsize{30.40\%} &
\normalsize{94.36\%}
\\

\normalsize{Spectral} & \normalsize{43.75\%} & \normalsize{43.37\%} & \normalsize{45.60\%} & \normalsize{93.66\%} & \normalsize{82.22\%}  & {\color[HTML]{000000} \normalsize{96.78\%}} &
\normalsize{50.00\%} &
\normalsize{43.37\%} &
\normalsize{40.00\%} &
\normalsize{93.66\%} \\

\normalsize{Elastic Subspace} & \normalsize{59.38\%} & \normalsize{49.40\%} & \normalsize{73.60\%} & \normalsize{94.37\%} & \normalsize{51.11\%}  & {\color[HTML]{000000} \normalsize{97.37\%}} &
\normalsize{48.43\%} &
\normalsize{40.96\%} &
\normalsize{52.00\%} &
\normalsize{94.37\%} \\

\normalsize{gMADD} & \normalsize{42.19\%} & \normalsize{46.99\%} & \normalsize{42.40\%} & \normalsize{72.54\%} & \normalsize{51.11\%}  & {\color[HTML]{000000} \normalsize{67.25\%}} &
\normalsize{39.06\%} &
\normalsize{44.58\%} &
\normalsize{35.20\%} &
\normalsize{58.45\%} \\

{\color[HTML]{000000} \normalsize{HDCC}} & {\color[HTML]{000000} \normalsize{59.38\%}} & {\color[HTML]{000000} \normalsize{34.94\%}} & {\color[HTML]{000000} \normalsize{29.60\%}} & {\color[HTML]{000000} \normalsize{50.00\%}} & {\color[HTML]{000000} \normalsize{40.00\%}}  & {\color[HTML]{000000} \normalsize{88.01\%}} &
\normalsize{51.50\%} &
\normalsize{38.55\%} &
\normalsize{29.60\%} &
\normalsize{50.00\%} \\

{\color[HTML]{000000} \normalsize{Leiden}} & {\color[HTML]{000000} \normalsize{50.00\%}} & {\color[HTML]{000000} \normalsize{46.99\%}} & {\color[HTML]{000000} \normalsize{68.00\%}} & {\color[HTML]{000000} \normalsize{71.12\%}} & {\color[HTML]{000000} \normalsize{82.22\%}}  & {\color[HTML]{000000} \normalsize{40.06\%}} &
\normalsize{48.43\%} &
\normalsize{46.99\%} &
\normalsize{49.60\%} &
\normalsize{71.13\%} \\

{\color[HTML]{000000} \normalsize{Louvain}} & {\color[HTML]{000000} \normalsize{42.19\%}} & {\color[HTML]{000000} \normalsize{48.19\%}} & {\color[HTML]{000000} \normalsize{76.00\%}} & {\color[HTML]{000000} \normalsize{94.34\%}} & {\color[HTML]{000000} \normalsize{82.22\%}}  & {\color[HTML]{000000} \normalsize{65.20\%}} &
\normalsize{45.31\%} &
\normalsize{48.19\%} &
\normalsize{60.80\%} &
\normalsize{93.66\%} \\

{\color[HTML]{000000} \normalsize{DP-GMM}} & {\color[HTML]{000000} \normalsize{46.88\%}} & {\color[HTML]{000000} \normalsize{43.37\%}} & {\color[HTML]{000000} \normalsize{54.40\%}} & {\color[HTML]{000000} \normalsize{85.92\%}} & {\color[HTML]{000000} \normalsize{73.33\%}}  & {\color[HTML]{000000} \normalsize{68.42\%}} &
\normalsize{45.31\%} &
\normalsize{44.58\%} &
\normalsize{39.20\%} &
\normalsize{92.96\%} \\

h{\color[HTML]{000000} \normalsize{CARP}} & {\color[HTML]{000000} \normalsize{43.75\%}} & {\color[HTML]{000000} \normalsize{46.99\%}} & {\color[HTML]{000000} \normalsize{36.00\%}} & {\color[HTML]{000000} \normalsize{75.25\%}} & {\color[HTML]{000000} \normalsize{71.11\%}} & {\color[HTML]{000000} \normalsize{79.82\%}} &
\normalsize{34.37\%} &
\normalsize{43.37\%} &
\normalsize{30.40\%} &
\normalsize{93.66\%}
\\

{\color[HTML]{000000} \normalsize{DEC}} & {\color[HTML]{000000} \normalsize{45.31\%}} & {\color[HTML]{000000} \normalsize{\textbf{71.08\%}}} & {\color[HTML]{000000} \normalsize{46.40\%}} & {\color[HTML]{000000} \normalsize{94.37\%}} & {\color[HTML]{000000} \normalsize{86.67\%}} & {\color[HTML]{000000} \normalsize{88.89\%}} &
\normalsize{54.69\%} &
\normalsize{\textbf{65.06}\%} &
\normalsize{33.60\%} &
\normalsize{94.37\%} \\

{\color[HTML]{000000} \normalsize{IDEC}} & {\color[HTML]{000000} \normalsize{48.44\%}} & {\color[HTML]{000000} \normalsize{67.47\%}} & {\color[HTML]{000000} \normalsize{61.60\%}} & {\color[HTML]{000000} \normalsize{92.96\%}} & {\color[HTML]{000000} \normalsize{73.33\%}} & {—}  & {—} & {—} & {—} & {—} \\

{\color[HTML]{000000} \normalsize{CarDEC}} & {\color[HTML]{000000} \normalsize{51.56\%}} & {\color[HTML]{000000} \normalsize{40.96\%}} & {\color[HTML]{000000} \normalsize{75.20\%}} & {\color[HTML]{000000} \normalsize{90.14\%}} & {\color[HTML]{000000} \normalsize{84.44\%}} & {—}  & {—} & {—} & {—} & {—} \\
\bottomrule
\end{tabular}
\endgroup
}
\label{tab:Table-2}
\end{table*}
\begin{table*}[!htb]
\centering
\captionsetup{justification=centering}
\caption{Clustering accuracy (ACC) and time elapsed (TOC) for consensus PCMF on large datasets. \\ 
(``X'' indicates computationally infeasible to run. ``T'' indicates infeasible due to run time out.)}
\resizebox{.88\columnwidth}{!}{%
\begingroup
\setlength{\tabcolsep}{4pt}
\begin{tabular}{@{}llccccccc@{}}
\toprule
 & \multicolumn{2}{c}{Tumors-Large; $N$ = 400} & \multicolumn{2}{c}{MNIST; $N$ = 36,000} & \multicolumn{2}{c}{Fashion MNIST; $N$ = 36,000} & \multicolumn{2}{c}{Synthetic; $N$ = 100,000} \\ 
 \midrule

 & \normalsize{ACC} & \normalsize{ACC (hold-out)} & \normalsize{ACC} & \normalsize{ACC (hold-out)} & \normalsize{ACC} & \normalsize{ACC (hold-out)} & \normalsize{ACC} & \normalsize{ACC (hold-out)} \\ 
 \midrule

\color[HTML]{F7890D} \normalsize{\textbf{PCMF}} & \textbf{{\color[HTML]{F7890D} \normalsize{100.00\%} }} & \color[HTML]{F7890D} \textbf{\normalsize{100.00\%}} & \textbf{\color[HTML]{F7890D} \normalsize{99.93\%}} & \color[HTML]{F7890D} \textbf{\normalsize{88.33\%}} & \textbf{{\color[HTML]{F7890D} \normalsize{99.94\%} }} & \textbf{{\color[HTML]{F7890D} \normalsize{81.41\%} }} & \textbf{{\color[HTML]{F7890D} \color[HTML]{F7890D} \textbf{\normalsize{100.00\%} }}} & \color[HTML]{F7890D} \textbf{\normalsize{100.00\%}} \\


\normalsize{PCA + K-means} & \normalsize{89.75\%} & \textbf{\normalsize{100.00\%}} & \normalsize{29.64\%} & \normalsize{29.64\%} & \normalsize{45.00\%} & \normalsize{45.48\%} & \normalsize{50.09\%} & \normalsize{50.00\%} \\

\normalsize{Ward} & \normalsize{90.50\%} & \normalsize{n/a} & \normalsize{--X--} & \normalsize{n/a} & \normalsize{--X--} & \normalsize{n/a} & \normalsize{--X--} & \normalsize{n/a} \\

\normalsize{Spectral} & \normalsize{92.00\%} & \normalsize{n/a} & \normalsize{--X--} & \normalsize{n/a} & \normalsize{--X--} & \normalsize{n/a} & \normalsize{--X--} & \normalsize{n/a} \\


\normalsize{gMADD} & \normalsize{61.50\%} & \normalsize{n/a} & \normalsize{--X--} & \normalsize{n/a} & \normalsize{--X--} & \normalsize{n/a} & \normalsize{--X--} & \normalsize{n/a} \\


\normalsize{Leiden} & \normalsize{66.25\%} & \normalsize{n/a} & \normalsize{60.62\%} & \normalsize{n/a} & \normalsize{38.31\%} & \normalsize{n/a} & \normalsize{10.88\%} & \normalsize{n/a} \\

\normalsize{Louvain} & \normalsize{72.25\%} & \normalsize{n/a} & \normalsize{69.88\%} & \normalsize{n/a} & \normalsize{42.26\%} & \normalsize{n/a} & \normalsize{10.85\%} & \normalsize{n/a} \\

\normalsize{DEC} & \normalsize{99.25\%} & \normalsize{n/a} & \normalsize{--T--} & \normalsize{n/a} & \normalsize{--T--} & \normalsize{n/a} & \normalsize{--T--} & \normalsize{n/a} \\

\normalsize{IDEC} & \normalsize{86.50\%} & \normalsize{n/a} & \normalsize{55.25\%} & \normalsize{n/a} & \normalsize{48.98\%} & \normalsize{n/a} & \normalsize{--T--} & \normalsize{n/a} \\

\bottomrule
\end{tabular}
\endgroup
}
\label{tab:Table-3}
\vspace{-1em}
\end{table*}

\textbf{PCMF reveals interpretable cluster dendrograms and predicts biomarkers.} First we evaluate our approach on 14 (7 single-view; 7 multiview) biomedical datasets and find it outperforms 14 clustering methods in nearly all cases (except against DEC/IDEC on SRBCT where it ranks 3rd; Table \ref{tab:Table-2}). 

Second, in the tumors-large dataset ($N=400$), we found that PCMF model coefficients for the F-Box And Leucine Rich Repeat Protein 2 (FBXL2) gene reveal a cluster hierarchy between GBM, lung, and breast cancer while a two-step approach does not (Fig. \ref{tumor-Figure}a-c). The branching structure reflects the suspected role of FBXL2 as a metastatic biomarker of breast-to-lung metastasis \citep{Wang2019-ws} and suggests a druggable target \citep{Deng2020-ls}. In Fig. \ref{tumor-Figure}d, Spearman's correlations between the PCMF score and prolactin receptor (PRLR) gene expression reveal strong slope differences between the three cancer tumors. PRLR is a mammary proto-oncogene \citep{Grible2021-ih,Sa-Nguanraksa2020-gk}, and a suggested prognostic biomarker of GBM progression (higher expression with shorter survival in males) \citep{Asad2020-za} and therapeutic target \citep{Asad2019-bm,Sa-Nguanraksa2020-gk}. Interestingly, PRLR is strongly but oppositely associated with the GBM ($R = -0.81$) and breast tumor clusters ($R = 0.45$), as suggested in literature on triple-negative breast cancer that shows higher expression is associated with lower recurrence and longer survival \citep{Motamedi2020-pq}.

Next, in a small $N$ COVID-19 dataset \citep{Shen2020-bs} we show that P3CA identifies hierarchical clustered metabolome-proteome embeddings that predict both severity ($ACC = 91.11\%$ in Table \ref{tab:Table-2}) and potential biomarkers. Severity hierarchy is not reflected in the two-step approach; Fig. \ref{COVID-Figure}a-b. Cluster-specific P3CA score Spearman's correlations with Carboxypeptidase B2 (CPB2) and Apolipoprotein M  (APOM) proteins (Fig. \ref{COVID-Figure}c-d) show opposite relationships---CPB2 (a known predictor of severe illness and a therapeutic target \citep{Claesen2022-cf, Foley2015-qf, Zhang2021-yf}) is strongly associated with the severe cluster ($R = -.8$). APOM (known to associate with less severity and better prognosis \citep{Cosgriff2022-zs,Shen2020-bs}) is only strongly associated with the not severe P3CA cluster ($R = 0.69$). Protein-protein interaction (PPI) networks constructed using the top 25 cluster-associated proteins reveal only a small PPI network for the healthy cluster with just 5/25 COVID-19-related genes versus large, highly-connected PPI networks with 15-18 COVID-19-related genes in the others (Fig. \ref{COVID-Figure}e-h; see methods description in Appendix $\S G$).

Finally, we demonstrate P3CA's utility for discovering autism subtypes, which could inform personalized diagnosis and therapies \citep{Buch2023-hw,Drysdale2017-il,Grosenick2019-iw} (Table 1 and Appendix $\S B2.2$). We find strong differences in the associations of autism \citep{Martino2014-wu,Martino2017-kk} subtype embeddings with behavior and brain connectivity (Appendix Fig. 7 and Appendix Tables 3--4), consistent with known autism subpopulation differences on behaviors with prefrontal cortex to somatosensory cortex, posterior parietal cortex, and middle temporal gyrus \citep{Buch2023-hw}. Subject-level P3CA embedding coefficients are robust to data perturbation (cosine similarity: $0.93 \pm 0.05$ for $U$ estimates and $0.97 \pm 0.03$ for $V$ estimates; 10 subsamples).

\section{Discussion and conclusion}
AI-enabled precision medicine promises dramatic improvements in healthcare, but facilitating adoption by healthcare professionals will require explainable, sensitive, and scalable methods appropriate for biomedical data. To meet this need, we have introduced a simple and interpretable joint clustering and embedding strategy using a modular convex clustering penalty. We instantiate our approach in three scalable algorithms that solve linear (PCMF), nonlinear (LL-PCMF), and multiview (P3CA) problems, and show that our method dominates standard convex clustering for $p > N$ data in the LDL regime. Empirically, our results on 14 biomedical datasets and 22 synthetic datasets demonstrate PCMF, LL-PCMF, and P3CA are highly competitive against 14 classical and state-of-the art (SOTA) clustering approaches. Further, our methods have superior explainability to SOTA clustering approaches as they enable the discovery of an interpretable hierarchy of cluster-wise embeddings that can predict diagnosis- and prognosis-relevant biomarkers. Still they have important \textbf{limitations}: in particular, they are less flexible than neural network methods, and are therefore likely to be dominated by such approaches in observation-rich cases. Overall, we present a simple and effective approach for healthcare professionals and neuroscientists to help customize biomarker discovery, diagnosis, prognosis, and treatment selection that is particularly effective in data-limited $p > N$ cases.

\bibliography{bib-arxiv2023}

\end{document}